\documentclass[10pt,twocolumn,letterpaper]{article}

\usepackage{cvpr}
\usepackage{times}
\usepackage{epsfig}
\usepackage{graphicx}
\usepackage{amsmath}
\usepackage{amssymb}
\usepackage{multirow}
\usepackage{xspace}
\usepackage{authblk}

\usepackage[breaklinks=true,bookmarks=false]{hyperref}
\cvprfinalcopy 

\def\cvprPaperID{****} 

\DeclareMathOperator{\softmax}{softmax}
\DeclareMathOperator{\entropy}{Entropy}
\DeclareMathOperator{\SPICE}{SPICE}

\newcommand{\SentiAttend}{\textsc{Senti-Attend}\xspace}
\newcommand{\SentiCap}{\textsc{SentiCap}\xspace}
\newcommand{\DirInj}{\textsc{Direct Injection}\xspace}
\newcommand{\SentFlow}{\textsc{Sentiment Flow}\xspace}
\newcommand{\Attend}{\textsc{Attend}\xspace}
\newcommand{\SentiAttendmEOneETwoL}{\textsc{Senti-Attend$_{-E_1E_2L_2}$}\xspace}
\newcommand{\SentiAttendmETwoL}{\textsc{Senti-Attend$_{-E_2L_2}$}\xspace}
\newcommand{\SentiAttendmL}{\textsc{Senti-Attend$_{-L_2}$}\xspace}

\begin{document}

\title{\SentiAttend: Image Captioning using Sentiment and Attention}

\author[1,2]{Omid Mohamad Nezami}
\author[1]{Mark Dras}
\author[2]{Stephen Wan}
\author[2]{C\'ecile Paris}
\affil[1]{Macquarie University,
Sydney, NSW, Australia}
\affil[2]{CSIRO Data61, Sydney, NSW, Australia}
\affil[ ]{\textit {\tt\small omid.mohamad-nezami@hdr.mq.edu.au, mark.dras@mq.edu.au \{stephen.wan,cecile.paris\}@data61.csiro.au}}


\maketitle

\begin{abstract}
   There has been much recent work on image captioning models that describe the factual aspects of an image. Recently, some models have incorporated non-factual aspects into the captions, such as sentiment or style. However, such models typically have difficulty in balancing the semantic aspects of the image and the non-factual dimensions of the caption; in addition, it can be observed that humans may focus on different aspects of an image depending on the chosen sentiment or style of the caption. To address this, we design an attention-based model to better add sentiment to image captions. The model embeds and learns sentiment with respect to image-caption data, and uses both high-level and word-level sentiment information during the learning process. The model outperforms the state-of-the-art work in image captioning with sentiment using standard evaluation metrics. An analysis of generated captions also shows that our model does this by a better selection of the sentiment-bearing adjectives and adjective-noun pairs.
\end{abstract}

\section{Introduction}

Image captioning systems aim to describe the content of an image using techniques from both computer vision and Natural Language Processing. The recent progress in designing visual object detection systems and language models together with large annotated datasets has led to a remarkable improvement in the ability to describe objects and their relationship displayed in an image. Generating factual captions is a challenging task as the image captioning systems are required to understand the semantic relation between the image and the caption. 

\begin{figure}[t]
    \begin{center}
    \includegraphics[width=0.50\linewidth]{COCO_val2014_000000290768.jpg}
    \end{center}
    \caption{An image with different foci for positive (``a beautiful well-appointed kitchen'') or negative (``ugly mugs'') sentiments.}
    \label{fig:kitchen}
\end{figure}

Moreover, when humans produce descriptions of images, they often go beyond the purely factual, and incorporate some subjective properties like sentiment or stylistic effects, depending on broader context or goals; a recent widely discussed example was the photo of Donald Trump at the 2018 G7 Summit where he is seated with arms crossed in front of Angela Merkel; various commentators have described the photo in negative (``eyes glaring'') or positive (``alpha male'') terms. Researchers in image captioning have similarly proposed models that allow the generation of captions with a particular style~\cite{gan2017stylenet} or sentiment~\cite{mathews2016senticap}. This incorporation of 
sentimental/emotional information has also been applied to text generation~\cite{hu2017toward} and conversation generation~\cite{zhou2017emotional}, and has been found to be an important element in image-grounded conversational agents~\cite{huber2018emotional}.

In sentiment-bearing image captioning, the focus of the caption could be different depending on the desired sentiment. Figure~\ref{fig:kitchen} contains an example image from the dataset of human-authored sentiment-infused captions from \cite{mathews2016senticap}, where all three negative captions focus on the mugs, and the positive captions focus on the light or the kitchen generally. Previous work in this domain usually does not use attention mechanisms~\cite{anderson2018bottom,lu2017knowing,rennie2017self,xu2015show}; however, the state-of-the-art image captioning models do, and the above observation suggests it would be useful here. The \SentiAttend model that we propose in this paper, therefore, incorporates attention into the generation of sentiment-bearing captions.

As a second observation, we note that previous work usually injected sentiment using one-hot vectors at each step of the caption generation~\cite{mathews2016senticap,you2018image}, which can have the effect of forcing sentiment into a generated description of an image that is not semantically suited to the image. The second contribution of \SentiAttend model, then, is to embed the sentiment information in real-valued vectors, allowing the model to learn where sentiment can be applied without changing semantic correlations between the image and the generated caption.
The model incorporates two kinds of sentiment embedding, which turn out to be complementary. The high-level embedding captures overall sentiment; it is fed into the long short-term memory (LSTM) network that handles the caption generation. The word-level embedding, in contrast, captures a notion of sentiment linked to the words in the vocabulary. The main contributions of the paper are highlighted as follows:

\begin{itemize}
    \item We propose an attention-based image captioning model called \SentiAttend to generate sentiment-bearing descriptions.
    
    \item Our model applies two complementary sentiment information including high-level and word-level. In addition to incorporating sentiments to the generated captions, this mechanism preserves the factual dimensions of the captions.
    
    \item The model outperforms the state-of-the-art models in generating sentimental descriptions.
\end{itemize}


\section{Related Work}
In our work we draw on both image captioning and controlled generation.

\subsection{Image Captioning}
Image captioning systems are usually designed according to a top-down paradigm and include a combination of a Convolutional Neural Network (CNN) and a long short-term memory (LSTM) to encode the visual content and generate the image descriptions, respectively~\cite{vinyals2015show}, which is inspired from the work of Sutskever \etal \cite{sutskever2014sequence}. The current state-of-the-art models in image captioning are attention-based systems~\cite{anderson2018bottom,lu2017knowing,rennie2017self,xu2015show}. The models use visual content, referred to as spatial features, as the input of an attention mechanism to selectively attend to different parts of an image at each time step in generating the image caption.

Yu \etal \cite{yu2017end} and You \etal \cite{you2016image} applied a notion of Semantic attention to detected visual attributes, which is learned in an end-to-end fashion. This attention model is used to attend to semantic concepts detected from various parts of a given image. Here, they used the visual content only in the initial time step. In other time steps, Semantic attention was used to select the extracted semantic concepts. That is, Semantic attention differs from spatial attention, which attends to spatial features in every time step, and does not preserve the spatial information of the detected concepts.

To preserve the spatial information, the salient regions are localized using spatial transformer networks~\cite{jaderberg2015spatial}, which get the spatial features as inputs. This is similar to Faster-RCNN generation of bounding boxes~\cite{ren2017faster}, but it is trained in an end-to-end fashion using bilinear
interpolation instead of a Region of interest pooling mechanism~\cite{johnson2016densecap}. Similarly, Anderson \etal \cite{anderson2018bottom} applied spatial features, but using a pre-trained Faster-RCNN and an attention mechanism to discriminate among different visual-based concepts regarding the spatial features.

\subsection{Controlled Generation}
Recently, Hu \etal \cite{hu2017toward} used variational autoencoders to control a generated sentence in terms of its attributes including sentiment and tense: they conditioned the sentence encoding process on these attributes. In conversation generation, Zhou \etal \cite{zhou2017emotional} used emotion categories to control the responses in terms of emotions. As a part of their system, they fed an embedded emotion as an input to their decoder. Ghosh \etal \cite{ghosh2017affect} proposed a model conditioning conversational text generation using affect categories. The model can control a generated sentence without previous knowledge about the words in the vocabulary. In our work, in contrast, we feed in embedded sentiments to capture both high-level and word-level sentiment information.

Moreover, image captioning systems control sentiment or other non-factual characteristics of the generated captions~\cite{gan2017stylenet,mathews2016senticap}. In addition to describing the visual content, these models learn to generate different forms of captions. For instance, Mathews \etal \cite{mathews2016senticap} proposed \SentiCap system to generate sentimental captions. Here, the notion of sentiment is drawn from Natural Language Processing~\cite{pang-lee:2008}, with sentiment either \textit{negative} or \textit{positive}. The \SentiCap system of Mathews \etal \cite{mathews2016senticap} is a full switching architecture incorporating both factual and sentimental caption paths. It needs two-stage training: training on factual image captions and training on sentimental image captions. Therefore, it does not support end-to-end training.

To address this issue, You \etal~\cite{you2018image} designed two new schemes, \DirInj and \SentFlow, to better employ sentiment in generating image captions. For \DirInj, an additional dimension was added to the input of a recurrent neural network (RNN) to express sentiment. A related idea was earlier proposed by Radford \etal~\cite{radford2017learning} who discovered a sentiment unit in a RNN-based system. In the \DirInj model, the sentiment unit is injected at every time step of the generation process. The \SentFlow approach of You \etal~\cite{you2018image} injects the sentiment unit only at the initial time step of a designated sentiment cell trained in a similar learning fashion to the memory cell in a long short-term memory (LSTM) network. Similar to the work of You \etal~\cite{you2018image}, we have a single phase optimization for our image captioning model. In contrast, \SentiCap, \DirInj and \SentFlow models apply visual features only in the initial time step of the LSTM.

However, recent state-of-the-art image captioning models usually apply visual features, which are spatial ones, at each time step of the LSTM~\cite{anderson2018bottom,lu2017knowing,rennie2017self,xu2015show}. Nezami \etal \cite{nezami2018face} also used spatial features at every step to generate more human-like captions, but for injecting facial expressions. In this work, therefore, we use an attention-based model to generate sentimental captions. In addition to this, we design a new injecting mechanism for sentiments to learn both high-level and word-level information. The high-level one involves injecting a sentiment vector to the LSTM. The fine-grained one involves using another sentiment vector to learn sentiment values appropriate at the word level.

\section{Approach}
Our image captioning model adapts a conventional attention-based encoder-decoder mechanism~\cite{anderson2018bottom,lu2017knowing,rennie2017self,xu2015show} to generate sentiment-bearing captions; we call our adapted model \SentiAttend.
Our model takes as the first input an image encoded into $K$ image feature sets, $a=\{a_{1},...,a_{K}\}$. Each set has $D$ dimensions to represent a region of the image, termed spatial features, $a_{i} \in \mathbb{R}^D$. $a$ is usually generated using a convolutional layer of a convolutional neural network (CNN). As the second input, we have the targeted sentiment category ($s$) to generate the image description with specific sentiment. The model takes these inputs and generates a caption
$x$ encoded as a sequence of 1-of-$K$ encoded words.

\begin{equation}
x = \{ x_1, \ldots, x_T \}, x_i \in \mathbb{R}^N
\label{equation:xt}
\end{equation}

\noindent
where $N$ is the size of the vocabulary and $T$ is the length
of the caption.

\subsection{Spatial Features}
ResNet-152~\cite{he2016deep} is used as the CNN model. It has been pretrained on the ImageNet dataset~\cite{simonyan2014very}. For use in our image captioning model, we use $7 \times 7 \times 2048$ features from the Res5c layer of the CNN model. Then, we reshape the features into $196 \times 512$ dimensions.

\subsection{Targeted Sentiment Category}
Our model aims to achieve an image description that is relevant to the targeted sentiment category. The sentiment categories are $\{$ Positive, Negative $\}$, as per the \SentiCap work~\cite{mathews2016senticap}. We have the further sentiment category $\{$ Neutral $\}$, which is for generating captions without dominant sentiment values. In our problem statement, we assume that the sentiment category is already specified, as previous work does. Because our model uses the targeted sentiment to describe an image, we can change the sentiment category to generate a different caption with a new sentiment value.

We embed the sentiment categories to give real value vectors, which are randomly initialized. Then, our model learns a sentiment category's vector during training time. Using this mechanism, we allow our system to learn the sentiment information in an adaptive way with respect to visual and text data. Specifically, we use one sentiment embedding vector ($E_1$) as an additional input to a long short-term memory (LSTM) network. In addition, we use another embedding vector ($E_2$) as a supplementary energy term to predict the next word's probability. $E_x$, the embedding vector of the sentiment category, has $F$ dimensions, $E_{x} \in \mathbb{R}^F$. $E_1$ is to model the high-level representation of the sentiment concept in the generated caption. $E_2$ represents desired word-level sentiment.

\subsection{Captioning Model}
In common with the architectures in \cite{anderson2018bottom,lu2017knowing,rennie2017self,xu2015show}, but with a sentiment term added,
\SentiAttend aims to minimize the following cross entropy loss (the first loss):

\vspace{-0.1cm}

\begin{multline}
L_1 = -\sum_{1 \leq t \leq T}\log( p_1(x_t \, | \, h_t, \hat{a}_t, E_2) ) +\\
\sum_{1 \leq k \leq K}(1-\sum_{1 \leq t \leq T}a_{tk})^2
\label{equation:loss1}
\end{multline} 

\begin{equation}
p_1 = {\softmax(h_{t}W_h + \hat{a}_{t}W_a + E_{2}W_e + b)}
\label{equation:w_prob}
\end{equation}

\vspace{-0.1cm}

\noindent
where $p_1$, as the categorical probability distribution across all words in the vocabulary, is from the output of a multilayer perceptron; $x_t$ is the next targeted word, is from the ground truth caption; $E_2$ is the second embedded vector of sentiment; $\hat{a}_t$ is the attention-based features, calculated using a conventional soft attention mechanism;
$h_t$ is the hidden state of the LSTM, estimated using Equation~\ref{equation:lstm}; and $W_x$ and $b$ are our trained weights and bias, respectively.
Similar to the previous work, the loss function contains a regularization term, to encourage
the system to take equal notice to different parts of
the image by the end of a caption generation process.

\SentiAttend also includes the following sentiment-specific cross entropy loss that we call the second loss:

\vspace{-0.1cm}

\begin{equation}
L_2=-\frac{1}{L}\sum_{1 \leq t \leq L}\log(p_2(s|h_t)) \quad
\label{equation:loss2}
\end{equation}

\begin{equation}
p_2 = {\softmax(h_{t}W_s + b_s)}
\label{equation:w_prob}
\end{equation}

\begin{figure}[t]
\begin{center}
\includegraphics[width=0.8\linewidth]{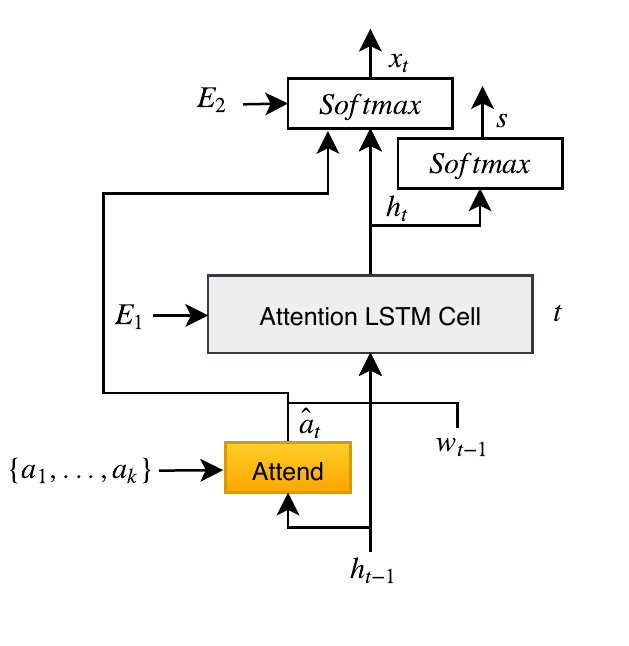}
\end{center}
\caption{The framework of the Senti-Attend model. Spatial image features $\{a_{1},...,a_{k}\}$ are attended using a conventional soft attention method. They are the outputs of our ResNet-152 network.\vspace{-0.25cm}}
\label{figure:senti-attend}
\end{figure}

\vspace{-0.1cm}

\noindent
where $p_2$ is the categorical probability distribution of the current state ($h_t$) across three sentiment classes $\{$ Positive, Neutral, Negative $\}$, obtained from a multilayer perceptron; $s$ is the targeted sentiment class; and $W_s$ and $b_s$ are our trained weights and bias. Using this loss function, \SentiAttend model can learn the targeted sentiment class at the end of each time step. The model architecture is shown in Figure~\ref{figure:senti-attend}.\footnote{We also tried optimizing the CIDEr metric~\cite{vedantam2015cider} similar to Rennie \etal~\cite{rennie2017self}; however, we could not achieve a better result. It is likely because we have a small number of captions with sentiment in our training set and optimizing the CIDEr metric leads to ignoring the sentiment-bearing parts of the captions.} 

Our LSTM decoder is defined by equations

\vspace{-0.4cm}

\begin{equation}
\begin{split}
& i_t = \sigma(W_{i}w_{t-1} + H_{i}h_{t-1} + A_{i}\hat{a}_{t} + B_{i}E_1 + b_i) \\
& g_t = \tanh(W_{g}w_{t-1} + H_{g}h_{t-1} + A_{g}\hat{a}_{t}  +B_{g}E_1 + b_g) \\
& o_t = \sigma(W_{o}w_{t-1} + H_{o}h_{t-1} + A_{o}\hat{a}_{t} + B_{o}E_1 + b_o) \\
& f_t = \sigma(W_{f}w_{t-1} + H_{f}h_{t-1} + A_{f}\hat{a}_{t} + B_{f}E_1 + b_f) \\
& c_t = f_{t}c_{t-1}+i_{t}g_t \\
& h_t = o_t\tanh(c_t) \quad
\end{split}
\label{equation:lstm}
\end{equation} 

\vspace{-0.3cm}

\noindent
where $i_t$, $g_t$, $o_t$, $f_t$ and $c_t$ are input gate, input modulation gate, output gate, forget gate and memory cell respectively. Here, $w_{t-1}$ is a real value vector with $M$ dimensions to represent the previous word, $w_{x} \in \mathbb{R}^M$, embedded using our model. $E_1$ is the first embedded vector of sentiment. It is used to condition the caption generation process using the targeted sentiment, which we refer as high-level sentiment. $W_x, H_x, A_x, B_x$, and $b_x$ are trained weights and biases and $\sigma$ is the logistic sigmoid activation function.

\begin{table*}
\begin{center}
\begin{tabular}{ |l| l| c| c| c| c| c| c| c| c| }
\hline
Senti & Model & B-1 & B-2 & B-3 & B-4 &  ROUGE-L & METEOR & CIDEr & SPICE \\ \hline\hline
\multirow{5}{*}{Pos}
& \Attend & 57.1 & 33.8 & 20.6 & 13.1 & 45.3 & 17.8 & 69.2 & 17.2 \\
& \SentiAttendmEOneETwoL & 56.0 & 32.3 & 18.7 & 11.1 &  43.6 & 18.7 & 60.8 & 15.9 \\
& \SentiAttendmETwoL & 57.1 & 34.1 & 21.0 & 13.2 &  45.5 & 18.1 & 69.9 & 16.8  \\
& \SentiAttendmL & 56.4 & 33.2 & 19.7 & 12.3 & 44.2 & 18.0 & 65.0 & 16.1 \\
& \SentiAttend  & 57.6 & 34.2 & 20.5 & 12.7 & 45.1 & 18.9 & 68.6 & 16.7  \\
 \hline
\multirow{5}{*}{Neg}
& \Attend & 56.5 & 33.5 & 20.2 & 12.5 & 45.0 & 17.7 & 67.7 & 16.3 \\
& \SentiAttendmEOneETwoL & 55.8 & 32.5 & 19.5 & 11.9 & 43.7 & 18.1 & 62.3 & 16.5 \\
& \SentiAttendmETwoL & 56.6 & 34.0 & 21.1 & 13.6 & 45.2 & 18.1 & 69.9 & 16.4 \\
& \SentiAttendmL & 56.6 & 34.2 & 21.5 & 14.1 & 45.4 & 18.0 & 71.3 & 16.8 \\
& \SentiAttend  & 58.6 & 35.4 & 22.3 & 14.7 & 45.7 & 19.0 & 71.9 & 17.4 \\
\hline
\multirow{5}{*}{Avg}
& \Attend & 56.80 & 33.65 & 20.40 & 12.80 & 45.15 & 17.75 & 68.45 & 16.75 \\
& \SentiAttendmEOneETwoL & 55.90 & 32.40 & 19.10 & 11.50 & 43.65 & 18.40 & 61.55 & 16.20 \\
& \SentiAttendmETwoL & 56.85 & 34.05 & 21.05 & 13.40 & 45.35 & 18.10 & 69.90 & 16.60\\
& \SentiAttendmL & 56.50 & 33.70 & 20.60 & 13.20 & 44.80 & 18.00 & 68.15 & 16.45\\
& \SentiAttend & \textbf{58.10} & \textbf{34.80} & \textbf{21.40} & \textbf{13.70} & \textbf{45.40} & \textbf{18.95} & \textbf{70.25} & \textbf{17.05} \\
\hline
\end{tabular}
\end{center}
\caption{The image captioning results (\%) of our models on the SentiCap test split. Pos, Neg, and Avg show the results on the positive test set, the negative test set and their average  (the best performances are bold). B-N is BLEU-N metric.}
\label{table:result_1}
\end{table*}

\begin{table*}
\begin{center}
\begin{tabular}{ |l| l| c| c| c| c| c| c| c| c| }
\hline
Senti & Model & B-1 & B-2 & B-3 & B-4 &  ROUGE-L & METEOR & CIDEr & SPICE \\ \hline\hline
\multirow{5}{*}{Pos}
& \SentiCap & 49.1 & 29.1 & 17.5 & 10.8 & 36.5 & 16.8 & 54.4 & \_ \\
& \DirInj & 51.2 & 30.6 & 18.8 & 11.6 & 38.4 & 17.2 & 61.1 & \_ \\
& \SentFlow & 51.1 & 31.4 & 19.4 & 12.3 & 38.6 & 16.9 & 60.8 & \_ \\
& Ours: \Attend & 57.1 & 33.8 & 20.6 & 13.1 & 45.3 & 17.8 & 69.2 & 17.2 \\
& Ours: \SentiAttend & 57.6 & 34.2 & 20.5 & 12.7 & 45.1 & 18.9 & 68.6 & 16.7 \\
 \hline
\multirow{5}{*}{Neg}
& \SentiCap & 50.0 & 31.2 & 20.3 & 13.1 & 37.9 & 16.8 & 61.8 & \_ \\
& \DirInj & 52.2 & 33.6 & 22.2 & 14.6 & 39.8 & 17.1 & 68.4 & \_ \\
& \SentFlow & 51.0 & 33.0 & 21.9 & 14.8 & 39.4 & 17.0 & 70.1 & \_ \\
& Ours: \Attend & 56.5 & 33.5 & 20.2 & 12.5 & 45.0 & 17.7 & 67.7 & 16.3 \\
& Ours: \SentiAttend & 58.6 & 35.4 & 22.3 & 14.7 & 45.7 & 19.0 & 71.9 & 17.4 \\
\hline
\multirow{6}{*}{Avg}
& SentiCap & 49.55 & 30.15 & 18.90 & 11.95 & 37.20 & 16.80 & 58.10 & \_ \\
& \DirInj & 51.70 & 32.10 & 20.50 & 13.10 & 39.10 & 17.15 & 64.75 & \_ \\
& \SentFlow & 51.05 & 32.20 & 20.65 & 13.55 & 39.00 & 16.95 & 65.49 & \_ \\
& Ours: \Attend & 56.80 & 33.65 & 20.40 & 12.80 & 45.15 & 17.75 & 68.45 & 16.75 \\
& Ours: \SentiAttend & \textbf{58.10} & \textbf{34.80} & \textbf{21.40} & \textbf{13.70} & \textbf{45.40} & \textbf{18.95} & \textbf{70.25} & \textbf{17.05}\\
& $\Delta$: Ours - Best Previous & \textbf{+6.40} & \textbf{+2.60} & \textbf{+0.75} & \textbf{+0.15} & \textbf{+6.30} & \textbf{+1.80} & \textbf{+4.76} &  \\
\hline
\end{tabular}
\end{center}
\caption{Our image captioning results (\%) compared to the state-of-the-art models on the SentiCap test split.}
\label{table:result_2}
\end{table*}

\section{Experimental Setup}
\subsection{Datasets}

\paragraph{Microsoft COCO dataset}
This dataset~\cite{chen2015microsoft}, which is the largest image captioning dataset, is used to train the \SentiAttend model. We use the specified training portion of the dataset which includes 413K+ captions for 82K+ images. This dataset can help our model to generate generic image captions.

\paragraph{SentiCap dataset}
This dataset~\cite{mathews2016senticap} is used to train our model to generate captions with sentiment. It includes manually generated captions with \textit{positive} and \textit{negative} sentiments. The training portion of the dataset consists of 2,380 positive captions for 824 images and 2,039 negative ones for 823 images.

For training our model, we combine the training sets of the Microsoft COCO and the SentiCap datasets similar to You \etal~\cite{you2018image}, where all captions in the Microsoft COCO dataset are assigned the \textit{neutral} label in terms of sentiment values. The validation set of the SentiCap dataset includes 409 positive captions for 174 images and 429 negative ones for 174 images. We use this set for model selection. Then the selected model is evaluated on the test set of the SentiCap dataset included 2,019 positive sentences for 673 images and 1,509 negative ones for 503 images. We separately report our results on the negative and positive parts of the test set similar to other work in this domain \cite{mathews2016senticap,you2018image}.

\subsection{Evaluation Metrics}
\label{sec:eval}
We evaluate \SentiAttend model using SPICE~\cite{anderson2016spice}, CIDEr~\cite{vedantam2015cider}, METEOR~\cite{denkowski2014meteor}, ROUGE-L~\cite{lin2004rouge}, and BLEU~\cite{papineni2002bleu}, which are standard evaluation metrics. Larger values are better for all metrics. Recently, the SPICE metric has shown a close correlation with human judgments. We have reported our results using SPICE for future comparisons, even though the previous work in this domain has not reported SPICE.

Moreover, to analyze the sentiment value of the generated captions using our model, all generated adjectives are tagged using the Stanford part-of-speech tagger software \cite{toutanova2003feature}. We apply the adjectives with strong sentiment values which are found in the list of the adjective-noun pairs (ANPs) of the SentiCap dataset (for example, \textit{cuddly}, \textit{sunny}, \textit{shy} and \textit{dirty}). Using Equation~\ref{equation:entropy}, we calculate Entropy. The variability of lexical selection can be indicated using Entropy (more variations corresponds to higher scores).

\begin{equation}
\entropy = -\sum_{1 \leq j \leq U}{\log_2(p(A_j))} \times p(A_j) \quad
\label{equation:entropy}
\end{equation} 

\noindent
where $A$ is an adjective and $p(A_j)$ is the probability of the adjective. $U$ is the number of unique generated adjectives.

We also calculate how often a model's chosen nouns correspond to the constituent nouns of the ANPs in the reference captions. To do this we implement a simplified version of SPICE that only considers noun matches, $\SPICE_N$; this includes the SPICE functionality of matching WordNet synonyms \cite{anderson2016spice} (for example, \textit{street} and \textit{road}). $\SPICE_N$ is an indication of how well the
model's captions attend to objects chosen by humans to add
a particular sentiment to.

\subsection{Models for Comparison}
To evaluate our model's performance in generating  sentiment-bearing captions, we compare it with the state-of-the-art approaches in this domain. Mathews \etal \cite{mathews2016senticap} introduced a joint-training model, which is trained on the Microsoft COCO dataset and then on the SentiCap dataset; this is the \SentiCap model. Subsequently, You \etal~\cite{you2018image} proposed two models for incorporating sentiment information into image captioning tasks, and they reported enhanced performance on the SentiCap dataset. The first of the two models, \DirInj, adds a new dimension to the input of their caption generator to express sentiment. In the second model, \SentFlow, they developed a new architecture to flow the sentiment information in different caption generation steps.

For model-internal comparisons, we use our attention-based image captioning model without sentiment inputs (the \Attend model); this allows us to assess the model's performance without sentiment. We further define three variants of our general Senti-Attend system (Figure~\ref{figure:senti-attend}). First, we use one-hot embedding representations for the ternary sentiment instead of the two embedding vectors ($E_1$ and $E_2$). We also do not use the second loss function. We call this model \SentiAttendmEOneETwoL. We speculate that the distributed representation of sentiment, analogous to those used in \cite{zhou2017emotional,ghosh2017affect}, will allow a model to be more selective about where to apply sentiment, where this one-hot variant in contrast might force sentiment inappropriately.
Second, we only use the first embedding vector ($E_1$) instead of both embedding vectors. Again, we do not use the second loss function. We call this approach the \SentiAttendmETwoL model. Third, we only leave out the second loss function, giving the \SentiAttendmL model. We name our full approach, which uses both embedding vectors and the second loss function, the \SentiAttend model. All Senti-Attend models have a similar architecture applying high-level and word-level sentiment information.

\subsection{Implementation Details}
In this work, we set the size of the memory cell and the hidden state of the LSTM to 2048 dimensions for all models except the \Attend model. The model has the memory cell and the hidden state with size of 1024 dimensions. The word embedding and the sentiment embedding vectors have 512 and 256 dimensions, respectively. The Adam optimization function with a learning rate of $0.001$ is applied to optimize our network~\cite{kingma2014adam}. We set the size of mini-batches to 180 for all models except the \Attend model, which has mini-batches of size 100. Since we have a small number of sentiment-bearing captions in our training set, we use dropout~\cite{srivastava2014dropout} on $E_1$ and $E_2$ to prevent overfitting. The size of our vocabulary is fixed to 9703 for all models. Since METEOR is more closely correlated with human judgments and is calculated more quickly than SPICE (METEOR does not need dependency parsing) \cite{anderson2016spice}, it is used to select the best model on the validation set.

\section{Results}

\subsection{Overall Metrics}
Table~\ref{table:result_1} shows the comparison between our models. The \SentiAttend model has achieved the best performances for all evaluation metrics across positive, negative and average test sets. The \SentiAttendmL and \SentiAttendmETwoL models have comparable performance (although we observe in Section~\ref{sec:res-qual} below that \SentiAttendmL produces better sentimental captions by those criteria). These two models, which are using the sentiment embedding vectors, outperform \SentiAttendmEOneETwoL for all evaluation metrics except METEOR, showing the effectiveness of our embedding approach. Here, we report the results of the \Attend model to show the performance of our attention-based system without sentiment; however, the model is not effective in terms of sentimental captions (Table~\ref{table:result_entropy}).

In Table~\ref{table:result_2}, we report the results of \SentiAttend compared to the state-of-the-art approaches on the SentiCap dataset. For a fair
comparison, our model is trained using a similar portion of Microsoft COCO and SentiCap datasets. We also use similar validation and test sets to evaluate our model. The model achieved better results compared to the previous state-of-the-art approaches by all standard evaluation metrics except Blue-4 for the negative test set, where the model is very marginally ($0.1$) lower than the \SentFlow model. However, the average results show that our model outperforms all state-of-the-art approaches on the SentiCap dataset. For example, our model achieved $6.40$, $6.30$, and $4.76$ improvements using Blue-1, ROUGE-L, and CIDEr metrics in comparison with the best previous state-of-the-art approaches, respectively. Using METEOR, the model achieved $1.80$ better performance compared to the approaches, which is showing closer correlation with human judgments. We also have $2.60$, $0.75$, and $0.15$ improvements using Blue-2 to Blue-4, respectively.

\begin{table*}

\begin{center}
\begin{tabular}{ |l| l| c| c| c| c| c| }
\hline
Senti & Model & $ER$ & $\SPICE_N$ & $C_{ANP}$  \\ \hline\hline
\multirow{5}{*}{Pos}
& \Attend & 2.2042 &  15.8 & 3 / 5 \\
& \SentiAttendmEOneETwoL & 3.2840 &  13.7 & 169 / 432 \\
& \SentiAttendmETwoL & 3.2795 &  15.3 & 8 / 25  \\
& \SentiAttendmL & 3.2691 & 14.4 & 60 / 171  \\
& \SentiAttend  & 3.2040  & 15.0 & 140 / 356\\
 \hline
\multirow{5}{*}{Neg}
& \Attend & 2.1513  & 15.5 & 2 / 5\\
& \SentiAttendmEOneETwoL & 3.5681  & 18.1 & 141 / 305 \\
& \SentiAttendmETwoL & 3.5895 & 16.7 & 11 / 29  \\
& \SentiAttendmL & 3.5396  & 17.7 & 35 / 76 \\
& \SentiAttend  & 3.7954  & 17.7 & 132 / 257\\
\hline
\multirow{5}{*}{Avg}
& \Attend & 2.17775  & 15.65 & 2.5 / 5.0  \\
& \SentiAttendmEOneETwoL & 3.42605  & 15.90 & 155.0 / 368.5 \\
& \SentiAttendmETwoL & 3.43450  & 16.00 & 9.5 / 27.0 \\
& \SentiAttendmL & 3.40435  & 16.05 & 47.5 / 123.5  \\
& \SentiAttend & \textbf{3.49970}  & \textbf{16.35} & 136.0 / 306.5 \\
\hline
\end{tabular}
\end{center}
\caption{The Entropy of the generated adjectives ($ER$) and the SPICE of the generated nouns ($\SPICE_N$) using our models on the SentiCap test split. We also calculated the number of the generated ANPs, both overall and only those that are in the reference captions ($C_{ANP}$).}
\label{table:result_entropy}
\end{table*}

\begin{table*}
\begin{center}
\begin{tabular}{ |l| l| l| }
\hline
Senti & Model & Top 10 Adjective \\ \hline\hline
\multirow{6}{*}{Pos}
& \Attend & white, black, small, blue, little, tall, different, \_, \_, \_ \\
& \SentiAttendmEOneETwoL & nice, beautiful, great, happy, good, busy, white, sunny, blue, black   \\
& \SentiAttendmETwoL & white, black, blue, small, nice, little, beautiful, tall, sunny, happy   \\
& \SentiAttendmL & great, nice, beautiful, white, black, blue, busy, right, healthy, happy  \\
& \SentiAttend & beautiful, nice, sunny, great, busy, white, blue, good, happy, calm \\
 \hline
\multirow{6}{*}{Neg}
& \Attend & white, black, small, blue, little, tall, different, busy, \_, \_   \\
& \SentiAttendmEOneETwoL & dirty, stupid, lonely, bad, broken, cold, dead, little, crazy, white \\
& \SentiAttendmETwoL & white, black, blue, small, little, dead, tall, busy, lonely, dirty \\
& \SentiAttendmL & dirty, white, black, dead, lonely, stupid, little, blue, broken, cold \\
& \SentiAttend & lonely, stupid, dead, dirty, broken, white, cold, black, bad, weird  \\
\hline
\end{tabular}
\end{center}
\caption{Top-10 generated adjectives using our models.}
\label{table:result_adj}
\end{table*}

\begin{figure*}
\begin{center}
\includegraphics[width=0.95\linewidth]{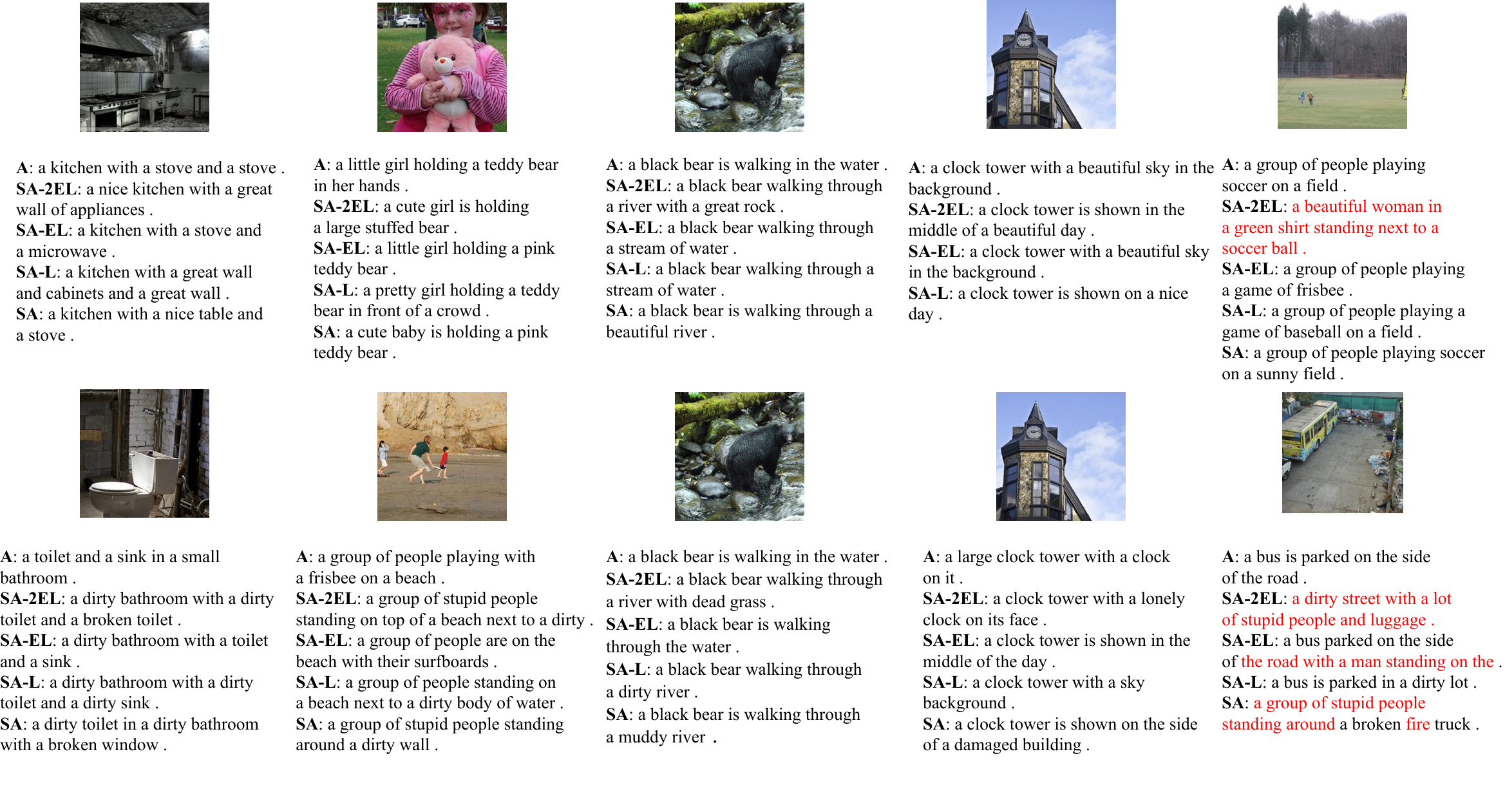}
\end{center}
\caption{Examples of generated captions using our models. The top row contains positive generated captions and the bottom row contains negative ones (A for \Attend, SA-2EL for \SentiAttendmEOneETwoL, SA-EL for \SentiAttendmETwoL, SA-L for \SentiAttendmL, and SA for \SentiAttend). The last column includes some captions with some inconsistent parts.}
\label{figure:samples}
\end{figure*}

\subsection{Qualitative Results}
\label{sec:res-qual}

In addition to calculating Entropy and $\SPICE_N$ explained in Section~\ref{sec:eval}, we calculate the number of ANPs generated by our models, both overall and only ANPs which are found in the (sentiment-appropriate) reference captions ($C_{ANP}$). From Table~\ref{table:result_entropy}, the \Attend model, which does not have sentiment inputs, achieves the lowest Entropy and $\SPICE_N$; moreover, in absolute numbers, it can only generate a few ANPs. The \SentiAttend model gives the highest Entropy and $\SPICE_N$, and so has both the highest variability in adjective choice for describing sentiment, and also attends best to the objects (represented by nouns) that human annotators choose to apply sentiment to in captions; in addition, it is not only accurate, it generates a large number of relevant ANPs. 

Although \SentiAttendmEOneETwoL can generate more ANPs, $C_{ANP}$ shows that \SentiAttend has better precision (measured by the ratio of the two terms) in generating ANPs which are compatible with the reference captions. $\SPICE_N$ further shows that \SentiAttendmEOneETwoL is less effective than other \SentiAttend models at choosing appropriate nouns. This accords with our intuition that the one-hot sentiment encoding do indeed force sentiment into captions in ways that are not optimal.
Of the remaining models, \SentiAttendmL can generate more ANPs. It also has the best $\SPICE_N$ after \SentiAttend. \SentiAttendmETwoL has the best Entropy after \SentiAttend and a good value of $\SPICE_N$; however, it does not generate many ANPs at all, indicating that having separate word-level sentiment is useful. As noted in the previous section, it achieved comparable performance to \SentiAttendmEOneETwoL for all overall evaluation metrics. This shows that generating both good performances for evaluation metrics (i.e. producing good captions in general), producing captions that have a high variety of sentiment terms and good alignment with objects chosen to apply sentiment to is a challenging issue in this domain. Nevertheless, the full \SentiAttend model is able to do well on all these competing objectives .


By way of examples, Table~\ref{table:result_adj} shows the top-10 generated adjectives using our models. We observe that colours are repeated for both positive and negative generated captions (as they are in both positive and negative ground-truth captions): these commonly appear in non-sentiment-based captioning system outputs. Among our sentiment models, \SentiAttendmETwoL ranks them high, whereas the others have sentiment-appropriate adjectives ranked higher. \textit{busy} is seen in positive generated captions more than the negative ones. It also has a higher rank in the positive ground-truth captions compared to the negative ones. \Attend has similar generated adjectives for both positive and negative ones because it does not use any sentiment signal.

Figure \ref{figure:samples} shows a number of the example generated captions using our models. The first two columns include positive and negative captions for different images. For example, in the first column, \SentiAttend generates the caption ``a dirty toilet in a dirty bathroom with a broken window''. The caption is compatible with the negative sentiment of the corresponding image. Other Senti-Attend models are also successful in generating negative captions for the image, but with less variability of expression. For instance, \SentiAttendmEOneETwoL generates ``a dirty bathroom with a dirty toilet and a broken toilet''. \SentiAttend can generate ``a kitchen with a nice table and a stove'' caption, which positively describes its corresponding image. \SentiAttendmL and \SentiAttendmEOneETwoL can also generate positive captions. Here, \Attend cannot generate sentiment-bearing captions. In the second column, we have similar generated captions. Columns three and four includes positive and negative captions for similar images. The captions show that \SentiAttend can manipulate and control the sentiment value of the captions using the targeted sentiment. For example, in the third column, \SentiAttend generates ``a black bear is walking through
a beautiful river'' and ``a black bear is walking through a
muddy river'' for the positive and negative sentiments, respectively. \SentiAttendmEOneETwoL can generate both types of sentiments, as well. \SentiAttendmL can generate a negative caption properly. In the forth column, all Senti-Attend models are successful in generating positive captions. The \Attend model also generates a positive caption properly. Here, generating negative captions for the image is even challenging for humans. However, \SentiAttend and \SentiAttendmEOneETwoL models can effectively generate positive (``nice day'' and ``beautiful day'') and negative (``damaged building'' and ``lonely clock'') ANPs for the image. They choose different nouns for different sentiments which are compatible with the corresponding image. The last column shows some captions with some errors. For example, \SentiAttendmEOneETwoL generates ``a beautiful woman in a green shirt standing next to a soccer ball'', which is not even semantically compatible with the image. In addition to generating sentimental captions, the captions should be compatible with their corresponding images. \SentiAttend can handle both topics properly although it has some errors in the generated captions (see \textcolor{red}{supplementary material} for more examples showing the effectiveness of the \SentiAttend model in comparison with other models to generate sentiment-bearing image captions).

\section{Conclusion}
In this work, we propose an attention-based image captioning model with sentiment, called the \SentiAttend model. One novel feature of this model is applying the attention-based features to effectively generate sentimental image captions. We also design a mechanism to incorporate embedded sentiments in generating image captions. The mechanism learns both high-level, which is conditioning our caption generator in general, and word-level, which is impacting on the word prediction process, sentiment information. We implement the mechanism with one-hot sentiment representation instead of sentiment embedding which results in less effective captions showing the effectiveness of our embedding approach. The \SentiAttend model significantly outperforms state-of-the-art work in this domain using all standard image captioning evaluation metrics. The qualitative results demonstrate that the improved performance is due to selecting sentimental adjectives and adjective-noun pairs in the generated captions.

\small{
\bibliographystyle{ieee}
\bibliography{egbib}
}

\clearpage

\begin{figure*}
\begin{center}
\includegraphics[width=1.0\linewidth]{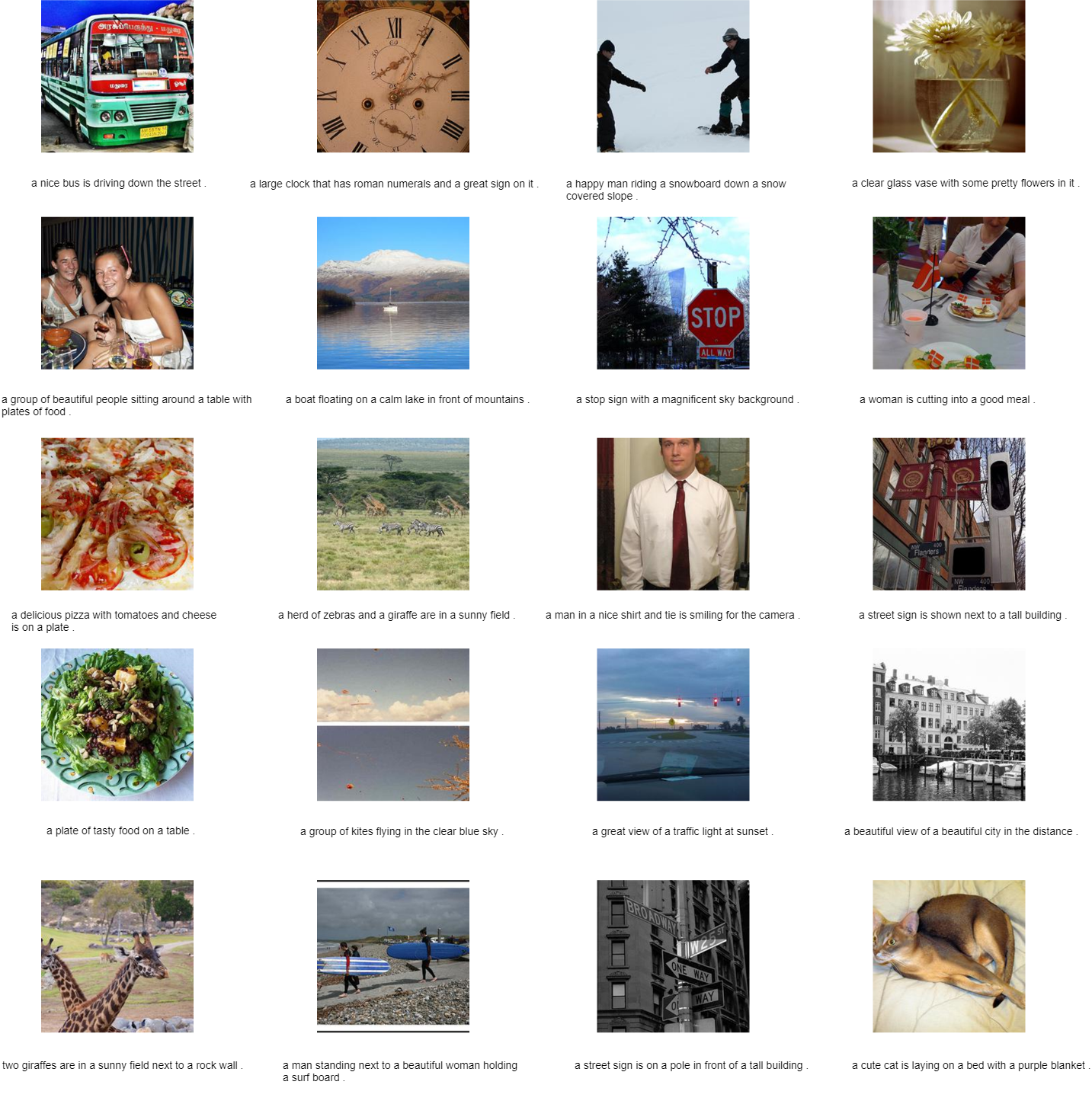}
\end{center}
\caption{Examples of positive generated captions with their corresponding images using the \SentiAttend model. The captions show the effectiveness of the model to generate positive captions which are compatible with the content of the images. For example, the model generates the relevant adjective-noun pairs (ANPs) including ``nice bus'', ``great sign'', ``happy man'', and ``pretty flowers'' in the first row, which are positively describing the images.}
\label{figure:pos_our_model}
\end{figure*}

\begin{figure*}
\begin{center}
\includegraphics[width=1.0\linewidth]{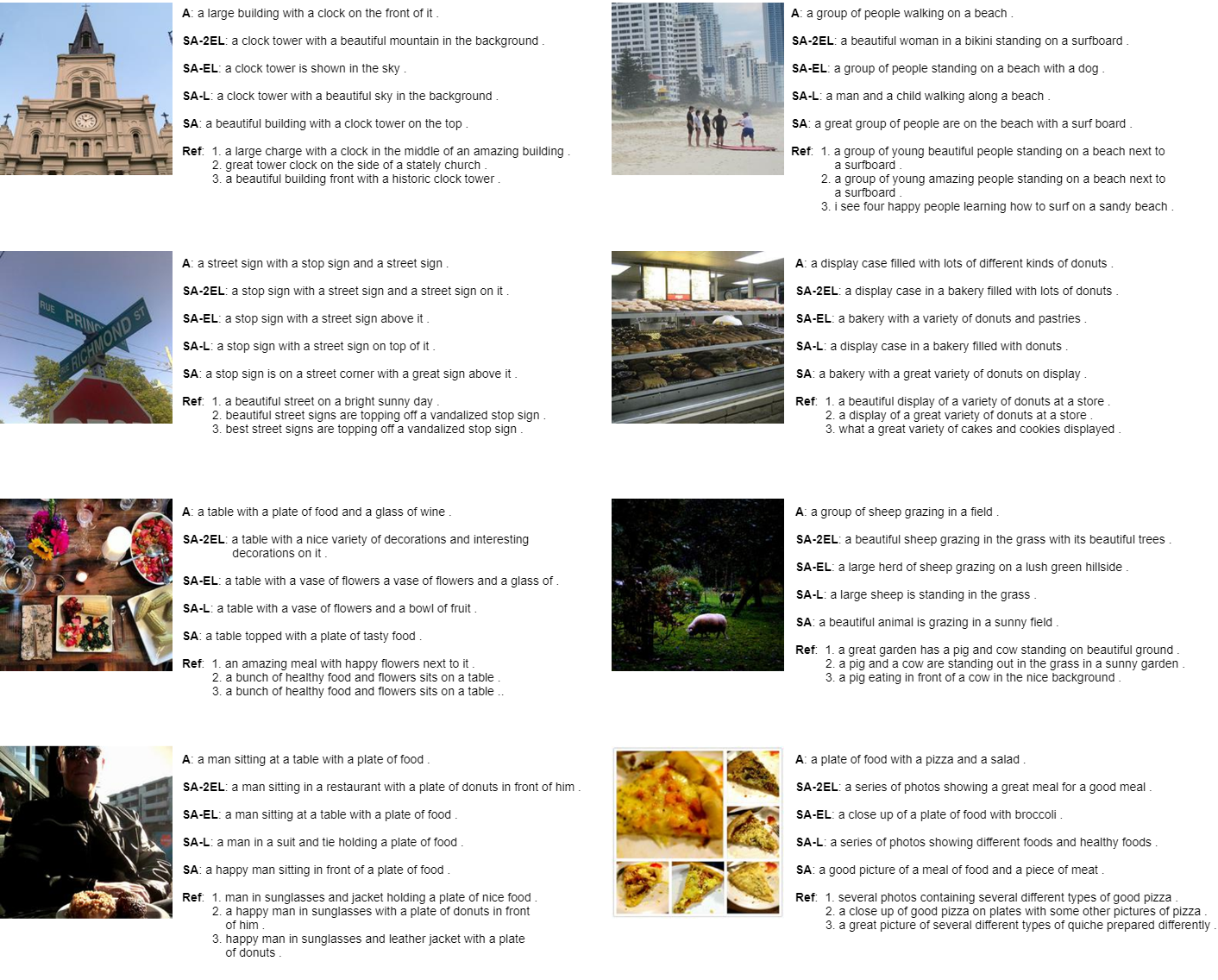}
\end{center}
\caption{Examples of positive generated captions using the \Attend (A), \SentiAttendmEOneETwoL (SA-2EL), \SentiAttendmETwoL (SA-EL), \SentiAttendmL (SA-L), and \SentiAttend (SA) models with their corresponding ground truth (Ref) descriptions. The SA model is the only approach which can add sentiment to the nouns having sentiment in the Ref descriptions. For instance, the model generates ``beautiful building'' in the first row where the Ref descriptions add ``amazing'' and ``beautiful'' to ``building''. It can also generate ``a great group of people'' in the first row. Here, in the Ref descriptions, we have ``a group of young beautiful people'' and ``a group of young amazing people'' which similarly add the positive sentiment to ``people''. }
\label{figure:pos_all_model}
\end{figure*}

\begin{figure*}
\begin{center}
\includegraphics[width=1.0\linewidth]{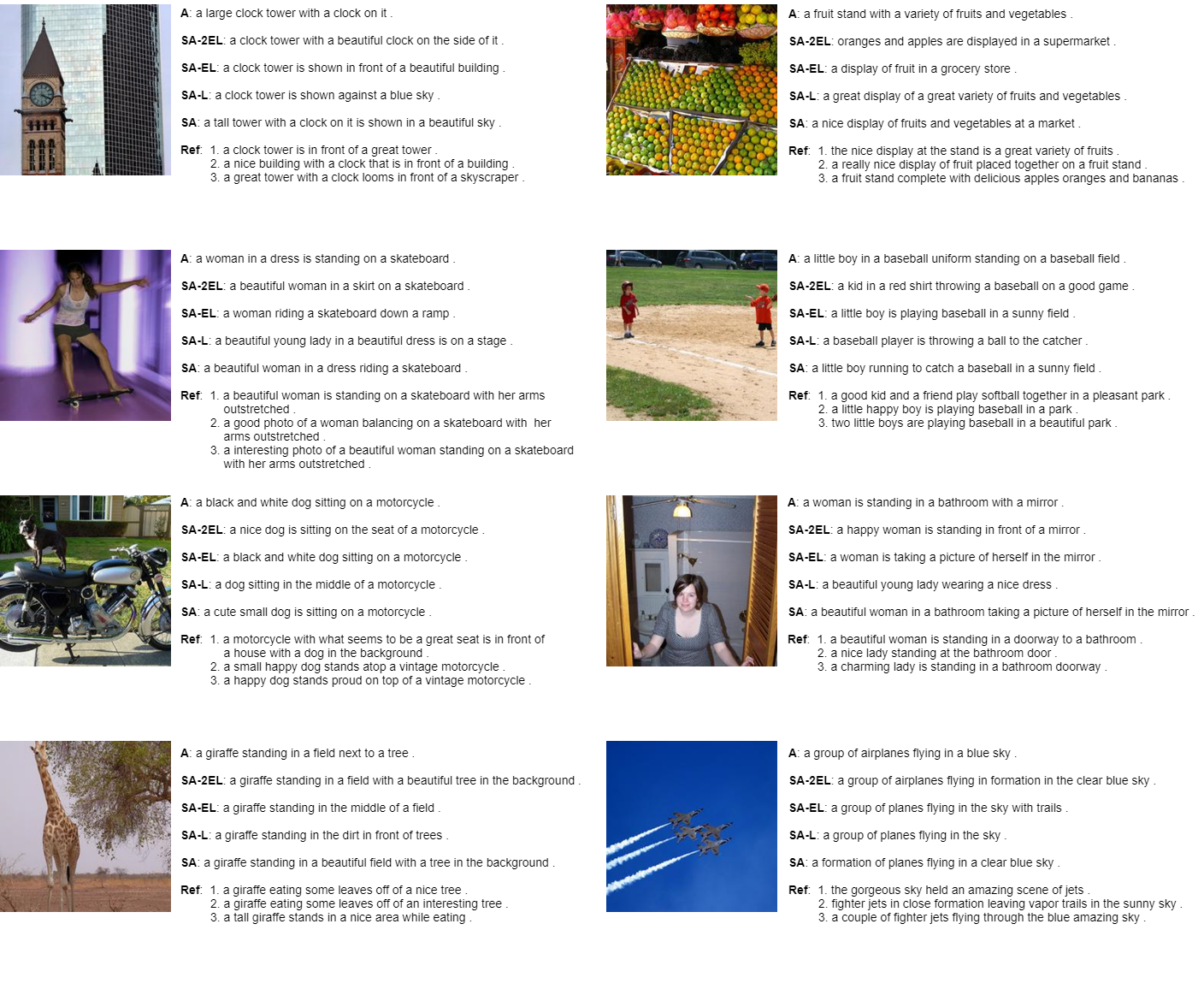}
\end{center}
\caption{Sample positive generated captions using all models with their Ref descriptions. Here, the models can add sentiment to the nouns having sentiment in the Ref descriptions; however, the SA model is the most successful one. For example, in the first row, it can generate ``tall tower'' where we have ``great tower'' in the Ref descriptions. Here, the SA-EL model also generates ``beautiful building'' where we have ``nice building'' in the Ref descriptions.}
\label{figure:pos_all_model}
\end{figure*}

\begin{figure*}
\begin{center}
\includegraphics[width=1.0\linewidth]{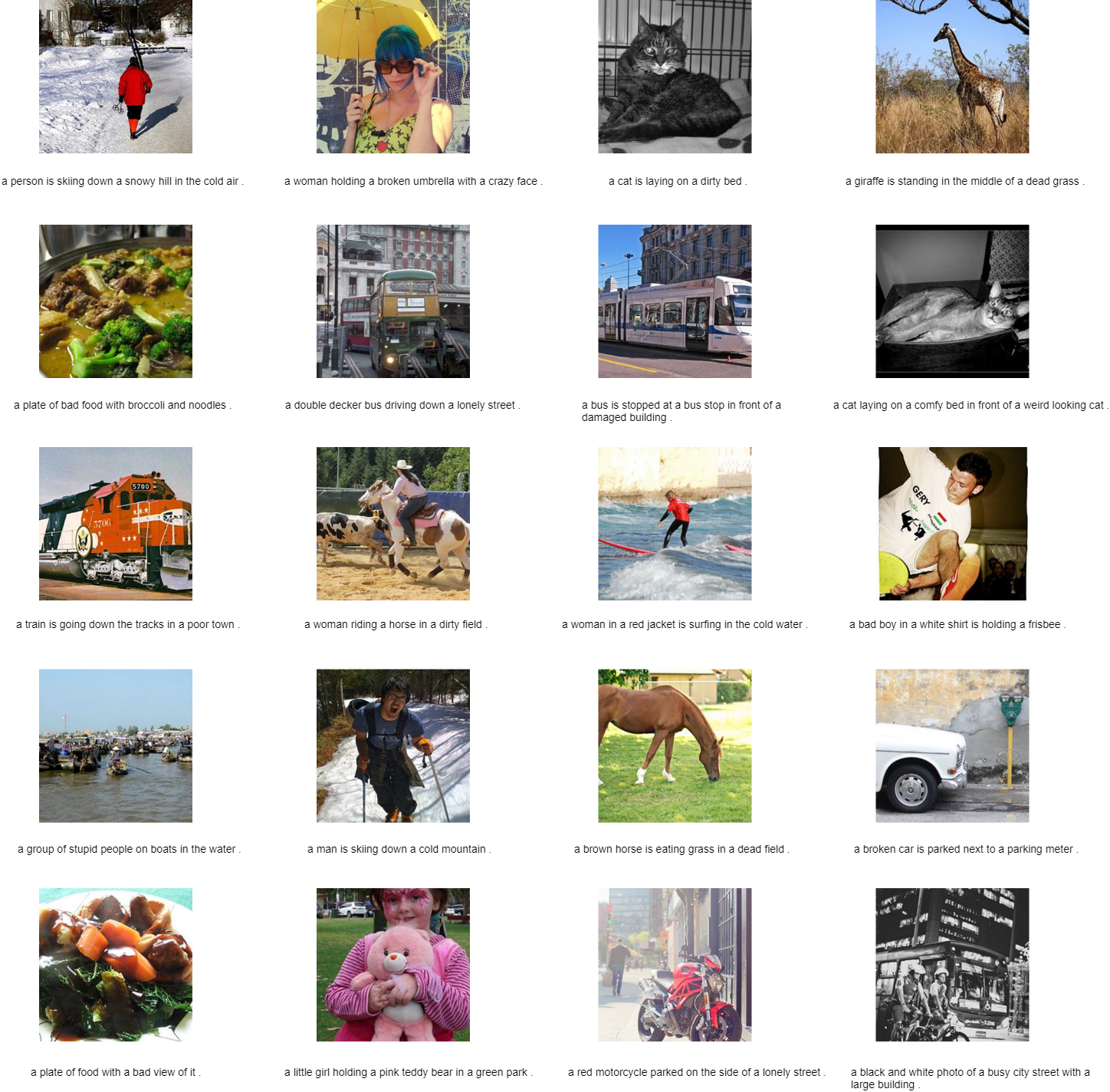}
\end{center}
\caption{Examples of negative generated captions using the \SentiAttend model, which are showing the effectiveness of the model to generate negative captions which are compatible with the visual content. For instance, the model properly generates the negative ANPs such as ``cold air'', ``broken umbrella'', ``crazy face'', ``dirty bed'', and ``dead grass'' in the first row. }
\label{figure:pos_all_model}
\end{figure*}

\begin{figure*}
\begin{center}
\includegraphics[width=1.0\linewidth]{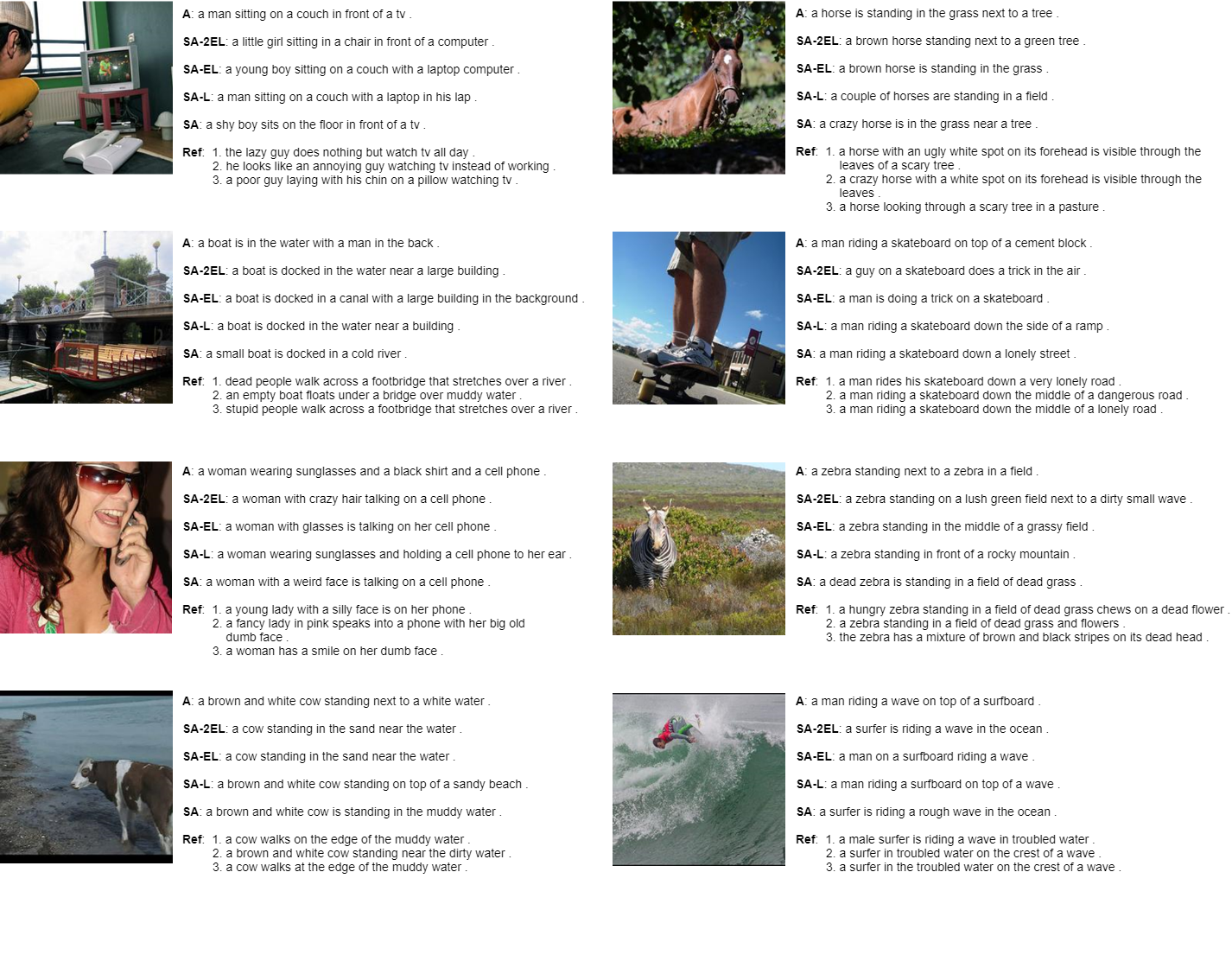}
\end{center}
\caption{Examples of negative generated captions using all models with their Ref descriptions. Only the SA model can add sentiment to its generated captions similar to the Ref descriptions. For example, it can generate ``shy boy''. Here, we have ``lazy guy'', ``annoying guy'', and ``poor guy'' in the Ref descriptions which are negatively describing the boy in the image. The SA-EL model can generate ``young boy''; however, ``young'' is not an adjective with a strong sentiment value in the SentiCap dataset.}
\label{figure:pos_all_model}
\end{figure*}

\begin{figure*}
\begin{center}
\includegraphics[width=1.0\linewidth]{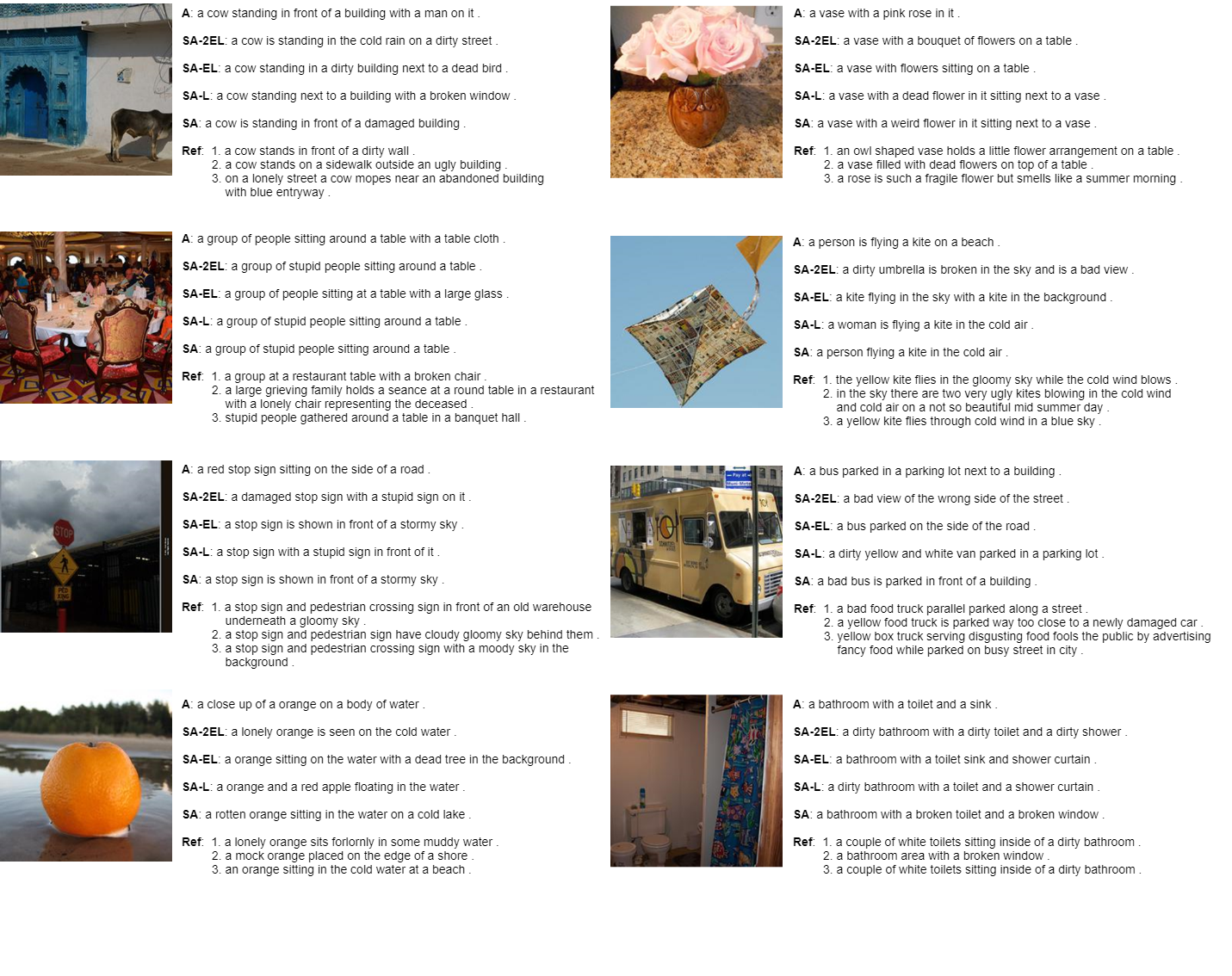}
\end{center}
\caption{Sample negative generated captions using the models with their corresponding Ref descriptions. Mostly, the SA model is successful to add sentiment to its generated captions which are similar to the Ref descriptions. For instance, in the first row, it can generate ``damaged building'' where we have ``ugly building'' and ``abandoned building'' in the Ref descriptions. ``building'', which is chosen by the SA model to add sentiment to, is a part of the Ref ANPs. In the first row, the SA model also generates ``weird flower'' where we have ``little flower'', ``dead flowers'' and ``fragile flower'' in the Ref descriptions. Here, the SA-L model correctly generates ``dead flower''. }
\label{figure:pos_all_model}
\end{figure*}

\end{document}